\documentclass[runningheads]{llncs}
\usepackage[T1]{fontenc}

\usepackage{graphicx}
\usepackage{amsmath}
\usepackage{adjustbox}    % for resize
\usepackage{array}        % improved array/tabular support
\usepackage{booktabs}     % nicer rules for tables

\newif\ifanonym
\anonymfalse   % <- set to \anonymtrue for submission (anonymous); \anonymfalse for camera-ready

\newcommand{\anonymize}[1]{%
  \ifanonym
    (anonymized)
  \else
    #1%
  \fi
}

\title{Using Text-Based Life Trajectories from Swedish Register Data to Predict Residential Mobility with Pretrained Transformers}
\titlerunning{Life Trajectories} % adjust or comment out if not needed

\usepackage{hyperref}
\ifanonym
  \hypersetup{pdftitle={Contribution Title},
              pdfauthor={},
              pdfsubject={},
              pdfkeywords={}}
\else
  \hypersetup{pdftitle={Contribution Title},
              pdfauthor={},
              pdfsubject={},
              pdfkeywords={}}
\fi

\ifanonym
  \author{Anonymous Authors}
  \authorrunning{Anonymous}
  \institute{} % leave empty for review
\else
  \author{Philipp Stark\inst{1,3}\orcidID{0000-0001-6529-1142} \and
          Alexandros Sopasakis\inst{2}\orcidID{0000-0001-9167-3590} \and
          Ola Hall\inst{1}\orcidID{0000-0002-9231-4028} \and
          Markus Grillitsch\inst{1,3}\orcidID{0000-0002-8406-4727}}
  \authorrunning{Stark et al.} % abbreviated list for the running head
  \institute{
    Department of Human Geography (KEG), Lund University, Sweden \and
    Department of Mathematics, Lund University, Sweden \and
    CIRCLE (Centre for Innovation, Research and Competence in the Learning Economy), Lund University, Sweden
  }
\fi

\begin{document}
\maketitle
\begin{abstract}
%150--250 words
We transform large-scale Swedish register data into textual life trajectories to address two long-standing challenges in data analysis: high cardinality of categorical variables and inconsistencies in coding schemes over time. Leveraging this uniquely comprehensive population register, we convert register data from 6.9 million individuals (2001–2013) into semantically rich texts and predict individuals’ residential mobility in later years (2013–2017). These life trajectories combine demographic information with annual changes in residence, work, education, income, and family circumstances, allowing us to assess how effectively such sequences support longitudinal prediction.
We compare multiple NLP architectures (including LSTM, DistilBERT, BERT, and Qwen) and find that sequential and transformer-based models capture temporal and semantic structure more effectively than baseline models. The results show that textualized register data preserves meaningful information about individual pathways and supports complex, scalable modeling. Because few countries maintain longitudinal microdata with comparable coverage and precision, this dataset enables analyses and methodological tests that would be difficult or impossible elsewhere, offering a rigorous testbed for developing and evaluating new sequence-modeling approaches. Overall, our findings demonstrate that combining semantically rich register data with modern language models can substantially advance longitudinal analysis in social sciences.

\end{abstract}

\section{Introduction}
Countries such as Sweden maintain exceptionally detailed administrative registers covering their entire populations over multiple decades\footnote{Statistics Sweden, \url{https://www.scb.se/en/}}. These linked records include information on demographics, residence, education, labor activity, or income, making them among the most comprehensive data sources in the world for studying social and economic life. Since few countries maintain longitudinal microdata with comparable precision and coverage, Swedish registers also enable method testing that is difficult or even impossible elsewhere, offering a rigorous testbed for developing and evaluating new methodological approaches. For the social sciences, Swedish registers serve a function similar to twin registries in medical research ~\cite{lichtenstein_swedish_2002}, enabling a class of questions that cannot be answered with any other data source. The richness and longitudinal nature of the data provide unique opportunities to examine how past pathways and decisions relate to future behavior and how individual behavior connects to societal dynamics ~\cite{woerz_social_2025,kroner_profiles_2025,dignum_empirically_2025,nilsson_self-initiated_2024,meyer_nationwide_2022}.

A distinctive characteristic of this data is that many key variables, such as education, occupation, or industry sectors, are categorical. These categories are defined by expert-designed taxonomies and represented through descriptive labels (e.g., “110: Manufacture of beverages”) organized in hierarchical classification systems. For instance, the Swedish industrial classification of 2007 contains 821 detailed industry categories that aggregate into 21 broader sectors~\cite{statistics_sweden_scb_mis_2007}. These systems are both highly detailed and semantically meaningful, as each category corresponds to a human-interpretable description.

Given their scale and granularity, such register data naturally invite the application of modern machine-learning methods that are capable of extracting high-dimensional patterns, capturing path dependencies, and modeling complex decision processes. However, two central challenges limit the direct use of categorical register data in machine-learning applications. First, categorical variables often have extremely high cardinality. Standard techniques such as one-hot encoding generate large, sparse matrices, amplifying the curse of dimensionality and reducing model efficiency ~\cite{liu_curse_2009,cerda_similarity_2018}. Second, coding schemes change over time. For example, the Swedish industrial classification was substantially revised in 2007, altering category boundaries and counts. This makes it difficult to harmonize longitudinal data and results in inconsistent variable spaces across years (see Table~\ref{tab:sni_mapping}).

To address these issues, we propose converting the category codes into their corresponding natural-language descriptions and representing the data directly as text. While this may appear counterintuitive from the perspective of traditional statistical modeling, which typically transforms real-world concepts into numeric representations, many register categories are semantically meaningful only through their textual definitions. Moreover, language model developments offer powerful tools for modeling semantic similarity and contextual meaning. Pretrained transformer models are expected to exploit the linguistic and hierarchical structure of these categories, making textual representations a promising way to circumvent both high cardinality and temporal coding inconsistencies.

\begin{table}[t]
\centering
\caption{Comparison of industry coding before and after the 2007 SNI revision.}
\scriptsize
\resizebox{0.85\textwidth}{!}{
\begin{tabular}{ll}
\hline
\textbf{Coding before 2007} & \textbf{Coding after 2007} \\
\hline
\textbf{Codes:} 64201, 64202, 64203 & \textbf{Code:} 61100 \\[2mm]
Network operation
    & Wired telecommunications activities \\
Radio and television broadcast operation
    & Wired telecommunications activities \\
Cable television operation
    & Wired telecommunications activities \\[2mm]
\hline
\textbf{Code:} 37100 & \textbf{Codes:} 38311, 38312, 38319, 38320 \\[2mm]
Recycling of metal waste and scrap
    & Dismantling of car wrecks \\
Recycling of metal waste and scrap
    & Dismantling of electric and electronic equipment \\
Recycling of metal waste and scrap
    & Dismantling of other wrecks \\
Recycling of metal waste and scrap
    & Recovery of sorted materials \\
\hline
\end{tabular}}
\label{tab:sni_mapping}
\end{table}

Based on this, we propose representing register data as textual life trajectories. Each trajectory begins with a static profile capturing demographic and socioeconomic characteristics and then unfolds through annual events such as residential moves, changes in work or education, income shifts, etc. Using these textualized sequences, we examine whether future residential mobility can be predicted from trajectories constructed with the data (2001–2013). We compare several Natural Language Processing (NLP) approaches (using Term Frequency–Inverse Document Frequency (TF–IDF) with linear classifiers to CNNs, LSTMs), encoder-only transformers (BERT), and compact Large Language Models (LLMs). This allows us to assess both the predictive value of semantically rich life trajectories and the effectiveness of different text-modeling architectures.

Our contributions are threefold:
(A) We introduce a scalable method for transforming large-scale register data into textual life trajectories that preserve semantic information and remain robust to coding inconsistencies.
(B) We demonstrate that these trajectories enable individual-level prediction of socially relevant outcomes, illustrated through residential mobility in Sweden.
(C) We compare frequency-based, convolutional, and transformer models and show that life trajectories contain predictive information beyond static demographics, providing a foundation for interpretable and scalable life-course modeling.

\section{Related Work}
Population register data provide detailed longitudinal information on individuals’ demographic, educational, occupational, health, and residential histories \cite{van_der_wel_gold_2019}. Such data have long supported life-course analyses linking work, education, family transitions, health, and mobility to socioeconomic outcomes \cite{dieleman_geography_2002,coulter_following_2013}. This has been demonstrated in research on dementia care trajectories \cite{zilling_formal_2025}, health outcomes across the life course \cite{keenan_health_2023}, and longitudinal crime patterns \cite{filser_are_2021}. The resolution and coverage of these registers enhance data analysis by capturing temporal ordering and long-term change \cite{ruspini_longitudinal_1999} and resonate with broader efforts to model individual trajectories and long-term socioeconomic outcomes \cite{yang_agent-based_2024}.

Residential mobility is a central life-course event involving complex interactions among family dynamics, employment, and socioeconomic stratification \cite{coulter_re-thinking_2016}. Research on GPS-based human mobility trajectories has shown the potential of embedding-based sequence representations \cite{li_understanding_2025}, and recent LLM-agent work has simulated daily personal movement patterns \cite{wang_large_2024}. These studies demonstrate that mobility can be framed as a sequence-modeling problem~\cite{sopasakis_traffic_2019}, though they focus on short-term travel behavior rather than multi-decade residential trajectories.

To address the high cardinality and temporal inconsistency of coding schemes in register data, prior research aggregated detailed categories \cite{mihaylov_measuring_2019,raisamo_s_predictors_2025} or restricted analyses to periods with consistent coding \cite{eriksson_industries_2013}. Consequently, existing machine-learning work on register data often predicts outcomes such as unemployment, income, or migration from manually engineered features derived from administrative codes \cite{jin_predicting_2023,noferesti_leveraging_2025}. This pattern also characterizes applied register-based research on mobility and residential transitions, such as studies of migrant selection and mortality \cite{korhonen_intermarriage_2025} or birth outcomes by migration status \cite{romero_intersectional_2025}. A recent approach treated Danish register data as structured sequences from which a BERT-based model could infer temporal and relational patterns \cite{savcisens_using_2024}. In this study, synthetic event tokens were constructed to represent occupational, health, and income transitions, enabling transformers to learn event embeddings. However, this framework relied on abstract event tokens rather than natural language and had not to address the changing coding schemes within their brief period (2008–2015).

The present study transformed register variables directly into natural-language life-trajectory narratives, allowing pretrained language models to operate on semantically meaningful descriptions. This representation preserved the hierarchical and descriptive content of occupational, industrial, educational, and geographical variables across coding revisions, enabling otherwise impossible long-period analysis. To our knowledge, no prior work has applied pretrained language models directly to such textualized register trajectories for predicting future social outcomes. 

\section{Method}

\subsection{Creating Life Trajectories and Labels}
\begin{table}[t]
\centering
\caption{Overview of variables, descriptions, and their corresponding change events used in the trajectory representation.}
\scriptsize
\begin{tabular}{lll}
\hline
\textbf{Variable} & \textbf{Description} & \textbf{Change Event} \\
\hline
Gender & Male / Female & -- \\
Age & Age in years & -- \\
Municipality & Place of residence & Residential move \\
Family Relation & Single, married, cohabiting, etc. & Family change \\
Children & Has children / Children grown up & Children status change \\
Education & Level and field of education & Education change \\
Employment & Employed, unemployed, outside labor force & Employment change \\
Occupation & SSYK (2001 or 2014) & Occupation change \\
Industry & SNI (2002 or 2007) & Industry change \\
Workplace Municipality & Municipality of workplace & Workplace move \\
Labor Market Region & Functional labor-market area & Labor-market move \\
Income & Main income source and income percentiles & Income change \\
Government Support & Receipt of government benefits & Change in government support \\
\hline
\end{tabular}
\label{tab:variables_events}
\end{table}

Individual life trajectories were constructed from annual Swedish register records (2001--2013) provided by the Longitudinal Integration Database for Health Insurance and Labor Market Studies (LISA)\footnote{\url{https://metadata.scb.se/mikrodataregister.aspx}}. We included individuals present in the registers before 2010 and still observable until 2015, ensuring at least four years of data for trajectory construction and at least two years for label computation, with 2013 as the split year. This selection yielded a sample of 6.99 million individuals. Each person’s baseline profile (sex, age, family, education, etc.) was combined with a sequence of annual life events (changes in family status, workplace, income, etc.). All variables and events are listed in Table~\ref{tab:variables_events}. Categorical codes were mapped to descriptive text using year-specific dictionaries. Table~\ref{tab:text} shows an example trajectory and illustrates how life trajectories were created from the register data. To compare static profiles against temporal event-based trajectories, we also constructed a dataset containing only individuals’ baseline profiles, excluding annual life events.

\begin{table}[t]
\caption{Illustration of the three-step transformation from coded SCB register variables to fluent life-trajectory text. First, numeric codes are looked up in code lists, then the resulting labels are combined into short natural-language fragments used as model input.}
\centering
\scriptsize
\begin{tabular}{p{0.32\linewidth} p{0.32\linewidth} p{0.32\linewidth}}
\hline
\textbf{Coded register variables (excerpt)} 
& \textbf{Intermediate labels after code lookup} 
& \textbf{Final life-trajectory text fragment} \\
\hline
Year = 2001; Sex = 1; Age = 34; ResMun = 138; \newline
FamilyRelation = 1; ChildStat = 0 &
Sex: Male \newline
Age: 34 years \newline
Home: Halmstad \newline
FamilyRel: Married \newline
Children: No children & In 2001 a male, aged 34, lives in Halmstad, is married and has no children. \\ \hline
EduLvl = 6; EduType = 343 & Education level: University degree \newline
Education type: Economics & The person has a university degree in economics. \\ \hline
Occupation = 4120; SNI = 6910 & Occupation: Financial assistant \newline 
Industry: Accounting and bookkeeping & The person works as a financial assistant in accounting and bookkeeping. \\ \hline
Year = 2004; ChildStat changes 0 $\rightarrow$ 1 & Children status changes: No children $\rightarrow$ Children &
In 2004, the person has children. \\ \hline
Year = 2006; ResMun changes 138 $\rightarrow$ 148 & Move: Halmstad $\rightarrow$ Göteborg & In 2006 the person moves from Halmstad to Göteborg. \\ \hline
\end{tabular}
\label{tab:text}
\end{table}

Residential mobility was calculated as a binary outcome indicating whether an individual changed municipality of residence at least once between 2013 and 2017 (proportion of movers = 0.136). The data was provided by Statistics Sweden under a restricted research license. Because they contain sensitive individual information, access is strictly regulated, and no microdata can be shared. All processing was carried out within the secure \anonymize{LUNARC \footnote{\url{https://www.lunarc.lu.se}}} high-performance computing facility at \anonymize{Lund University}. However, to ensure full reproducibility, all code and processing pipelines, as well as small synthetic data examples are provided in the \anonymize{\url{https://github.com/PhilippStark/TLT-PT}} repository.
 
\subsection{Dataset Preparation}
Life trajectories were randomly split, reserving $\approx$5\% as a held-out test set (350k samples). As an initial exploratory step, we embedded the test-set trajectories using the Qwen3 4B massive text embedding\footnote{\url{https://huggingface.co/Qwen/Qwen3-Embedding-4B}}
. We then applied PCA to reduce the 2560-dimensional embeddings to 50 components. Then, t-SNE projected them into two dimensions, allowing us to visualize part of the test samples and color-code them by their mobility label.

For model training, a TF–IDF vectorizer transformed the texts into weighted 1–2-gram features using sublinear term-frequency scaling, L2 normalization, and up to 300k features. For transformer models, the pretrained BERT or Qwen tokenizer encoded the trajectories into token IDs truncated to each model’s maximum input length. Because BERT supports fewer input tokens (maximum 512), we examined token lengths and found that 99\% of the trajectories contained fewer than 496 tokens.

\subsection{Model Selection and Training}
We compared different architectures: a TF–IDF logistic regression baseline~\cite{sammut_tfidf_2011}; CNN and LSTM models for local and sequential dependencies; the contextualized transformer models BERT~\cite{devlin_bert_2018} and DistilBERT~\cite{sanh_distilbert_2019}; and Qwen3–0.6B~\cite{yang_qwen3_2025} as a compact LLM. Besides one fully fine-tuned DistilBERT model, we applied a training strategy that updated only the \emph{embeddings + classification head}, leaving the transformer backbone frozen. Under this setup, DistilBERT trained 24.4M of its 67M parameters (36.5\%), including its pre-classifier layer. BERT trained 23.8M of 109.5M parameters (21.8\%), while Qwen3--0.6B trained 155.6M of 596.1M parameters (26.1\%). This strategy reduces computational cost while still allowing the model to adapt to the downstream task.

All models were trained on 1M randomly sampled life-trajectory texts (fixed seed for reproducibility). Preliminary experiments indicated that performance saturated before 500k samples, motivating this sample size choice. All token-based models were trained for two epochs using the AdamW optimizer (learning rate: $5\times10^{-5}$, weight decay: 0.01, warmup ratio: 0.05). For BERT and Qwen, we used the HuggingFace Trainer with automatic batch-size selection to remain within GPU memory limits. We addressed class imbalance using a weighted cross-entropy loss, and selected the best-performing checkpoint based on validation AUPRC. Unless stated otherwise, models were trained and evaluated using identical preprocessing, tokenization, and sampling setups. No hyperparameter tuning was performed. Evaluation was conducted on a held-out test set using balanced accuracy, macro-F1, per-class precision and recall, and AUPRC, where the baseline AUPRC corresponds to the positive-class prevalence of 0.136.

All models were trained once on an NVIDIA A100 GPU for approximately 160-GPU hours in total. Experiments were conducted in Python 3.12 with CUDA 12.1 using the packages \texttt{transformers} 4.57.1, \texttt{torch} 2.9.1, and \texttt{tokenizers} 0.22.1. 
Baseline models were implemented with \texttt{scikit-learn} 1.7.1 and \texttt{scipy} 1.16.1.

\section{Results}
The sociodemographic patterns observed between movers and non-movers are consistent with established findings: younger individuals show substantially higher mobility, while those with children are less likely to move (Table~\ref{tab:mobility_descriptives}). Movers also tend to have longer recorded life trajectories, likely reflecting a greater number of documented events.

\begin{table}[t]
\small
\centering
\caption{Sociodemographic and economic characteristics by residential mobility outcome. “Previous Mobility” refers to moves during 2001–2013. Tokens show the mean and standard deviation of tokenized life trajectories used as model input.}
\scriptsize
\begin{tabular}{lll}
\hline
\textbf{Variable} & \textbf{No Mobility} & \textbf{Mobility} \\
\hline
Female (\%) & 50.8 & 49.3 \\
Age (mean $\pm$ SD) & 40.4 $\pm$ 16.9 & 26.9 $\pm$ 14.2 \\
Higher Education (\%) & 65.4 & 40.6 \\
Single (\%) & 39.6 & 40.0 \\
No Children (\%) & 47.0 & 26.9 \\
Income (mean $\pm$ SD) & 47.2 $\pm$ 33.9 & 35.2 $\pm$ 31.2 \\
Previous Mobility (\%) & 24.1 & 56.9 \\
Tokens (mean $\pm$ SD) & 203.3 $\pm$ 95.4 & 251.9 $\pm$ 107.4 \\
\hline
\end{tabular}
\label{tab:mobility_descriptives}
\end{table}

The TF--IDF logistic regression baseline achieved a balanced accuracy of 0.732 and an AUPRC of 0.409, clearly exceeding the base-rate of movers (13.6\%) (see Table \ref{tab:main_results}). Neural models trained on tokenized life trajectories performed slightly better overall: TextCNN reached a balanced accuracy of 0.739 and an AUPRC of 0.429, and TextLSTM achieved 0.737 and 0.429, respectively. Among transformer-based models, full fine-tuning of DistilBERT yielded competitive results (BalAcc = 0.734; AUPRC = 0.417), whereas restricting training to the embeddings and classifier led to weaker performance (BalAcc = 0.715; AUPRC = 0.366). BERT with trainable embeddings and classifier reached a balanced accuracy of 0.730 and an AUPRC of 0.461. Qwen3--0.6B in the same configuration performed similarly on balanced accuracy (0.732) but achieved the highest AUPRC among the transformer variants (0.467). Models relying solely on static baseline profiles performed noticeably worse than trajectory-based models, with TextCNN achieving 0.700 balanced accuracy and 0.348 AUPRC.

\begin{table}[t]
\small
\centering
\caption{Test performance for all models. Metrics: Balanced Accuracy (BalAcc), AUPRC, F1-macro, and per-class precision/recall for movers (label = 1) and non-movers (label = 0). 
Groups: A – TF--IDF features; B – Tokenized texts; C – Tokenized baseline profiles. 
Proportion of movers: 13.6\%.}
\scriptsize
\resizebox{\textwidth}{!}{
\begin{tabular}{clccccccc}
\hline
\textbf{Group} & \textbf{Model (variant)} &
\textbf{BalAcc} & \textbf{AUPRC} &
\textbf{F1$_{\text{macro}}$} &
\textbf{Prec$_{1}$} & \textbf{Rec$_{1}$} &
\textbf{Prec$_{0}$} & \textbf{Rec$_{0}$} \\
\hline
A & Logistic Regression              & 0.732 & 0.409 & 0.631 & 0.306 & 0.721 & 0.944 & 0.744 \\
% "auroc": 0.800
\hline
B & TextCNN                          & 0.739 & 0.429 & 0.641 & 0.318 & 0.722 & 0.945 & 0.756 \\
%"eval_auroc": 0.809
B & TextLSTM                         & 0.737 & 0.429 & 0.640 & 0.317 & 0.715 & 0.944 & 0.758 \\
%"eval_auroc": 0.806
B & DistilBERT (embedding + classifier) & 0.715 & 0.366 & 0.625 & 0.300 & 0.677 & 0.937 & 0.752 \\
%'eval_auroc': 0.7742
B & DistilBERT (full fine-tuning)    & 0.734 & 0.417 & 0.636 & 0.313 & 0.716 & 0.944 & 0.753 \\
% "eval_auroc": 0.8007
B & BERT (embedding + classifier)    & 0.730 & 0.461 & 0.560 & 0.431 & 0.577 & 0.932 & 0.884 \\

B & Qwen3 0.6B (embedding + classifier) & 0.732 & 0.467 & 0.560 & 0.425 & 0.585 & 0.933 & 0.879 \\
\hline
C & TextCNN                          & 0.700 & 0.348 & 0.618 & 0.293 & 0.646 & 0.931 & 0.755 \\
\hline
\end{tabular}}
\label{tab:main_results}
\end{table}

\begin{figure}[t]
    \centering
    \includegraphics[width=0.8\textwidth]{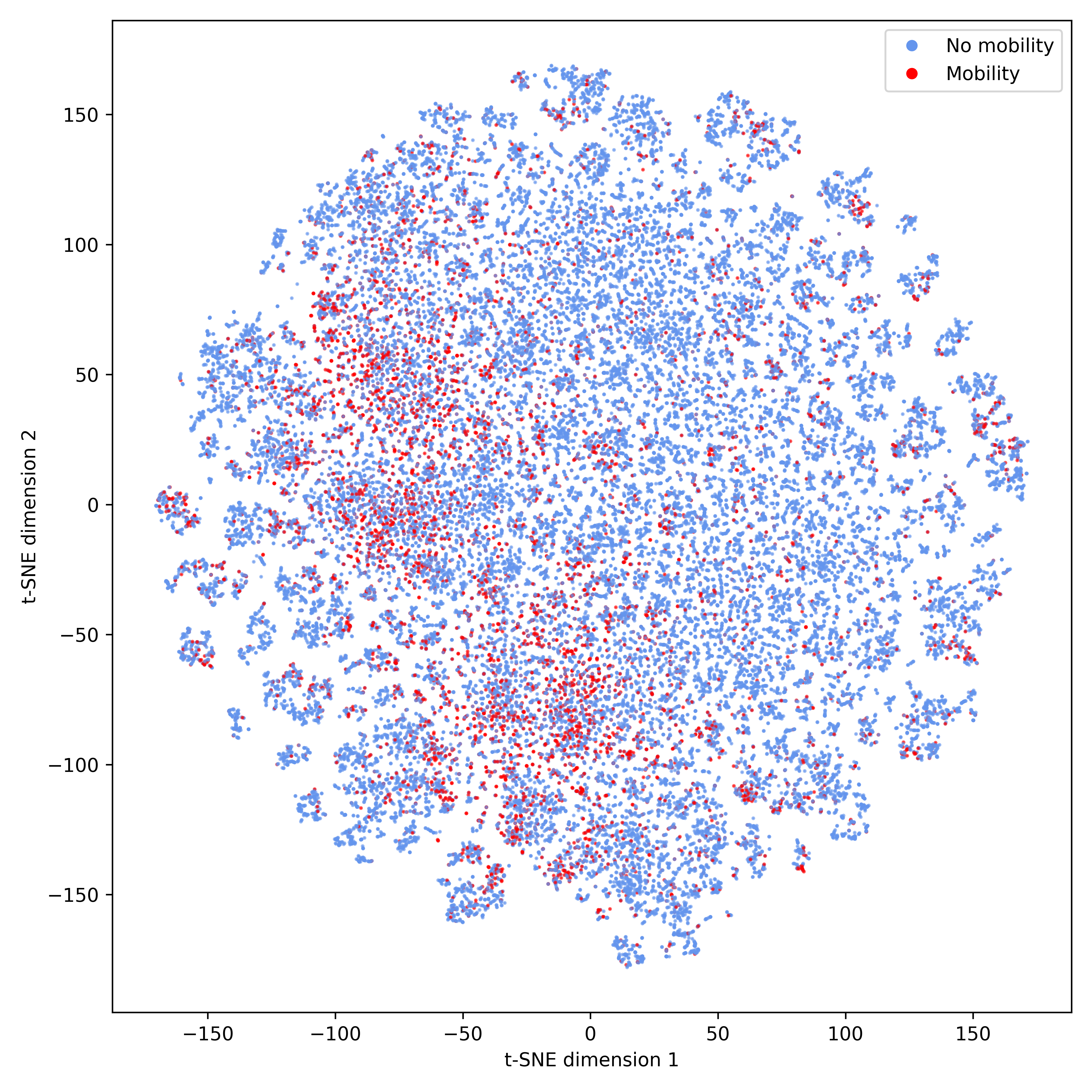}
    \caption{t-SNE visualization (perplexity: 10) of a random subsample of 50,000 life-trajectory 
    embeddings from the test set, created using the Qwen3 4B massive text embedding model. 
    Points are colored by residential mobility status (mobility vs.~no mobility).}
    \label{fig:tsne_trajectories_with_labels}
\end{figure}

For visualization purposes, a subset of the test data was used to generate two-dimensional t-SNE projections of the Qwen3--4B trajectory embeddings after reducing them to 50 principal components. We tested a range of perplexity values (10, 20, 50, 100, 200) and observed that, across all settings, mover trajectories were consistently concentrated in several of the larger clusters located below zero in both t-SNE dimensions. While the granularity and shape of clusters varied with the perplexity parameter, the overall distribution of movers relative to non-movers remained stable (see Fig \ref{fig:tsne_trajectories_with_labels}).

\section{Discussion}
This study examined whether residential mobility can be predicted from textualized life trajectories derived from Swedish register data. Despite substantial class imbalance, the results indicated that life trajectories contain meaningful information for distinguishing movers from non-movers. Static baseline profiles already captured some of the predictive signals. Adding longitudinal events provided additional value, suggesting that changes in work, education, family, or residence contribute incremental information. For non-movers, whose life courses are typically more stable, the static snapshot appeared more sufficient to represent the most relevant characteristics.

Looking at the t-SNE visualization, movers consistently appeared more concentrated in a specific subregion across all tested perplexity settings. This indicates that the embeddings capture local similarities between mover trajectories, even though movers occupy a diffuse rather than sharply separated area of the latent space. The absence of clear clusters from an untrained embedding combined with a simple unsupervised projection suggests that residential mobility does not follow a single homogeneous pattern but instead reflects multiple, diverse pathways.

Model behavior aligns with these structural properties. The low-capacity neural models show high recall but very low precision for the minority class. Under strong class imbalance, recall typically remains stable, while precision is sensitive to false positives, making it the first metric to deteriorate when models struggle to capture fine-grained distinctions~\cite{saito_precision-recall_2015}. These models, therefore, detect many potential movers but cannot reliably separate true from false positives. In contrast, the larger transformer models achieve substantially higher precision while maintaining similar recall, indicating improved discrimination between mover and non-mover trajectories. The consistent increase in AUPRC across transformer variants supports this interpretation by reflecting stronger early-retrieval quality and more informative ranking of individuals by mobility risk.

Overall, the results show that representing register data as textual life trajectories is a viable and promising approach. Even without domain-specific knowledge or fine-tuning, compact transformer models already achieve solid predictive performance, despite their size being far smaller than contemporary foundation models. The life-trajectory representation itself provides a scalable and harmonized format for longitudinal prediction and enables new analytical possibilities. The framework can be extended by enriching the models with domain knowledge or by leveraging representation learning from larger pretrained models. Importantly, the strong performance of the compact models demonstrates that small and efficient architectures can already be effective in developing tailored social-science applications. This makes them particularly suitable for explainable AI methods and hybrid modeling frameworks, where model transparency, computational feasibility, and interpretability are essential.

% Left out bvecause of space:
% First, the register data captures only individuals who remain in Sweden. Cross-border movers exit the dataset, which may bias the outcome. Second
At the same time, limitations must be acknowledged. First, administrative registers record formal events and may omit informal, subjective, or contextual influences on mobility. Second, the current models rely exclusively on individual-level sequences and do not exploit relational or geographic dependencies, such as ties to workplaces, partners, or local labor markets. Given that spatial reasoning remains a challenge even for larger language models~\cite{guidotti_text_2026}, the absence of explicit geographic structure likely constrains performance. Third, external macro-level factors (economic shocks, housing-market conditions, or policy changes) are not represented in the trajectories. Finally, we emphasize that the purpose of these models is to generate population-level insights that inform society and evidence-based policy making, not to evaluate, target, or classify individuals.

\section{Conclusion}
We showed that transforming register data into text-based life trajectories provides a practical solution to issues of high cardinality and coding inconsistencies and enables meaningful longitudinal modeling. The resulting trajectories proved effective when used with pretrained transformer models, including compact LLMs, demonstrating that rich, event-level register data can be successfully integrated into modern language-model architectures. These findings offer a promising starting point for combining the contextual depth and fine-grained detail of population registers with the representational power of contemporary language models, opening new opportunities for predictive modeling and exploratory analysis in the social sciences.

\ifanonym
    \begin{credits}
    \subsubsection{\ackname} Anonymous

    \subsubsection{\discintname} Anonymous
    \end{credits}

\else
    \begin{credits}
    \subsubsection{\ackname} This research was supported by the Swedish Research Council for Sustainable Development (FORMAS), grant no. 2022-151862.
    
    \subsubsection{\discintname}
    The authors have no competing interests.
    \end{credits}
\fi

\bibliographystyle{splncs04}
\bibliography{references.bib}
\end{document}